\makeatletter\def\graphicscache@inhibit{true}\makeatother

\documentclass[letterpaper, 10 pt, conference]{ieeeconf}  %
\usepackage[hidelinks]{hyperref}
\usepackage[style=ieee,natbib=true,backend=bibtex,firstinits,doi=false,%
     mincitenames=1,maxcitenames=2,maxbibnames=99,sorting=none,terseinits=false]{biblatex}
\bibliography{references.bib}

\usepackage{graphicscache}
\usepackage{tikz}
\usetikzlibrary{arrows}
\usetikzlibrary{positioning,calc}
\usetikzlibrary{decorations.pathreplacing}
\usetikzlibrary{decorations.markings}
\usetikzlibrary{fit}
\usetikzlibrary{shapes.callouts}
\usetikzlibrary{shapes.geometric}
\usetikzlibrary{matrix}

\usepackage{amsmath}
\usepackage{amssymb}
\usepackage{textcomp}
\usepackage{dsfont}

\usepackage[capitalize]{cleveref}

\usepackage[export]{adjustbox}

\usepackage{pgfplotstable}
\usepackage{booktabs}

\usepackage{placeins}

\usepackage[draft,nomargin,inline]{fixme}
\fxsetup{theme=color}

\usepackage{threeparttable}

\IEEEoverridecommandlockouts                              %

\title{\LARGE \bf
Refining 6D Object Pose Predictions\\ using Abstract Render-and-Compare
}

\author{Arul Selvam Periyasamy*, Max Schwarz*, and  Sven Behnke%
\thanks{*: The authors contributed equally.}%
\thanks{All authors are with the Autonomous Intelligent Systems (AIS) Group,
Computer Science Institute VI, University of Bonn, Germany.  }%
\thanks{Email: {\tt\small periyasa@ais.uni-bonn.de}}%
}

\begin{document}

\maketitle
\thispagestyle{empty}
\pagestyle{empty}

\begin{abstract}
Robotic systems often require precise scene analysis capabilities, especially
in unstructured, cluttered situations, as occurring in human-made environments.
While current deep-learning based methods yield good estimates of object poses,
they often struggle with large amounts of occlusion and do not take inter-object
effects into account.
Vision as inverse graphics is a promising concept for detailed scene
analysis. A key element for this idea is a method for inferring scene parameter
updates from the rasterized 2D scene. However, the rasterization process
is notoriously difficult to invert, both due to the projection and occlusion
process, but also due to secondary effects such as lighting or reflections.
We propose to remove the latter from the process by mapping the rasterized image
into an abstract feature space learned in a self-supervised way from pixel
correspondences.
Using only a light-weight inverse rendering module, this allows us to refine
6D object pose estimations in highly cluttered scenes by optimizing a simple
pixel-wise difference in the abstract image representation.
We evaluate our approach on the challenging YCB-Video dataset, where it
yields large improvements and demonstrates
a large basin of attraction towards the correct object poses.
\end{abstract}

\section{Introduction}
\label{sec:intro}

Robust robotic interaction in environments made for humans is an open research
field. An important prerequisite in this context is scene perception, yielding
the necessary information such as detected objects and their poses or affordances
for later manipulation actions.
While there are various high-accuracy methods for scene understanding,
the problem becomes significantly harder in the presence of clutter and
inter-object effects.
As such, current works in humanoid manipulation that require precise grasping
are often limited to non-cluttered
or even isolated scenes (e.g. \citep{pavlichenko2018autonomous,klamt2018supervised}).
While the manipulation action itself and planning for it is certainly more difficult
in cluttered scenes, robust 6D object pose estimation is a necessary prerequisite.

An interesting approach in this context is the idea of viewing computer vision
as an inverse graphics process~\cite{grenander1976lectures,grenander1978lectures}.
It promises to perform
scene analysis by inverting the rasterization process, which sounds highly
promising---today's rendering techniques are capable of producing convincing
photo-realistic renderings of highly complicated scenes, so inversion of the
process should yield high-quality scene analysis.
However, the problem plaguing the inverse graphics field is that the rendering
process is largely unidirectional, with complex physical effects such as lighting,
surface scattering, transparency, and so on. Furthermore, scene analysis is
especially in demand for cluttered scenes, e.g. in warehouse automation contexts,
but occlusion effects caused by clutter are among the most difficult to invert or differentiate.

To take a step towards a solution of this problem, we propose to first remove
most \textit{secondary} rendering effects from the scene using abstract
surface features learned in an unsupervised manner. This way, only the
\textit{primary} effects remain---occlusion and projection.
These effects can then be explained and analyzed by a simpler differentiable
rendering component.

We apply our render-and-compare framework to the task of monocular 6D pose
estimation, specifically pose refinement, where initial pose guesses are available.
In our approach, 6D pose predictions from state-of-the-art pose estimation methods are refined by
minimizing the pixelwise difference between the rendered image and the observed image 
in the proposed abstract descriptor space invariant to secondary rendering effects.
We further make the assumption that meshes of the objects are available, as
is the case in many industrial and robotic applications. For example,
service robots operating in human environments working with tools designed for human usage and
industrial part handling robots can greatly benefit from having precise 6D pose estimation. 

\begin{figure}
   \centering
   \scalebox{0.95}{%
\begin{tikzpicture}[
	fmap/.style={rectangle,draw=black,fill=yellow!20},
    font=\scriptsize\sffamily,
    label/.style={anchor=north,align=center, text depth=0},
	axislabel/.style={font=\footnotesize,inner sep=1pt},
]

\node[inner sep=0] (input) at (-4,0) {\includegraphics[width=.95\linewidth]{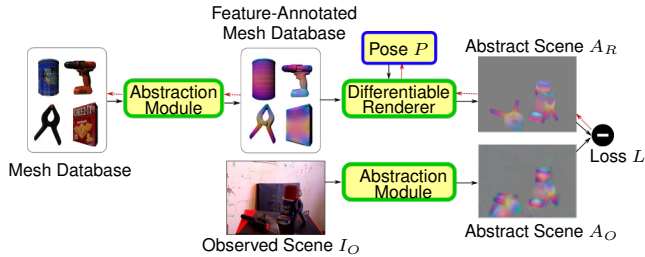}};

\node[label] at (-7.5, -0.3) {\scriptsize Mesh Database};
\node[label] at (-4.5, 1.9) {\scriptsize Feature-Annotated\\\scriptsize Mesh Database};
\node[label] at (-4.5, -1.35) {\scriptsize Observed Scene $I_O$ };

\node[label] at (-.9, 1.45) {\scriptsize Abstract Scene $A_R$};
\node[label] at (-.9, -1.15) {\scriptsize Abstract Scene $A_O$};

\node[label] at (-6.05, .8) {\scriptsize Abstraction};
\node[label] at (-6.05, .57) { \scriptsize Module};

\node[label] at (-2.9, 0.8) {\scriptsize Differentiable};
\node[label] at (-2.9, 0.58) {\scriptsize Renderer};

\node[label] at (-2.8, -.38) {\scriptsize Abstraction };
\node[label] at (-2.8, -.6) { \scriptsize Module };

\node[label] at (-2.9, 1.4) {\scriptsize Pose $P$};

\node[label] at (0.2, -0.1) {\scriptsize Loss $L$ } ;

\end{tikzpicture}
}
   \vspace{-.5cm}
   \caption{Scene analysis using differential rendering and learned abstraction
   module. An abstract representation is extracted both for the input scene
   and the mesh database.
   The differential rendering module then tries to match the abstract scene
   representation with the feature-annotated meshes.
   The loss gradient (red dotted line) only needs to
   be backpropagated through the differentiable renderer.}
   \label{fig:refinement:abstraction_render}
\end{figure}

In short, our contributions proposed in this work include:
\begin{enumerate}
 \item A scene abstraction method that removes secondary render effects, so
    that scene analysis by render-and-compare becomes feasible,
 \item fusion of surface features onto object meshes for direct rendering
    of scenes in the abstract feature space,
 \item a fast and light-weight differentiable rendering component, and
 \item the integration of these components into a pose refinement pipeline,
    which is evaluated on the YCB Video Dataset~\cite{xiang2017posecnn}.
    To the best of our knowledge, this is the first differentiable rendering
    pipeline capable of optimizing object poses in cluttered real-world scenes.
\end{enumerate}

This paper is structured as follows. After discussing related works in the section II, we
describe the descriptor learning process in section III and introduce our differentiable renderer, LightDR in section IV.
In section V, we apply our pipeline to 6D object pose refinement,
and further analyze the robustness of the proposed refinement process.

\section{Related Work}

Vision as inverse graphics aims at inferring object parameters like shape, illumination, reflectance, and pose, scene parameters like camera parameters, lighting, and secondary reflections by inverting the rendering process. The process of rendering 3D scene to discrete 2D pixels involves discretization steps that are not differentiable. However, several approximation methods have been to proposed to realize a differentiable renderer.  \citet{loper2014opendr} proposed OpenDR, a generic differentiable renderer that can compute gradients with respect to object and scene parameters. \citet{kato2018neural} introduced a differentiable renderer that is suited for neural networks. \citet{rezende2016unsupervised} treat forward rendering as a black-box and used REINFORCE~\cite{williams1992simple} to compute gradients. \citet{li2018differentiable} proposed edge sampling algorithm to differentiate ray tracing that can handle secondary effects such as shadows or global illumination. \citet{liu2019soft} proposed a differentiable probabilistic formulation instead of discrete rasterization. The differentiable renderer used in this work is closely modeled after OpenDR but tailored for object pose refinement with a strong focus on speed.

Vision as inverse graphics is most often formulated as a render-and-compare approach, where model parameters are optimized by minimizing the difference between rendered and observed images. \citet{zienkiewicz2016real} used render-and-compare for real-time height mapping fusion. Several recent works used render-and-compare for solving a wide range of vision problems: \citet{tewari2017mofa} learned unsupervised monocular face reconstruction; \citet{kundu20183d} introduced a framework for instance-level 3D scene understanding; \citet{moreno2016overcoming} estimated 6D object pose in cluttered synthetic scenes.
More closely related is the DeepIM method by \citet{li2018deepim}, who
formulated 6D object pose estimation as an iterative pose refinement process
that refines the initial pose by trying to match the rendered image with the observed image.
In contrast to our approach, they avoid the need for backpropagating gradients
through the renderer by training a neural network to output pose updates.
While the method yields very promising results, it is not directly clear how to
apply this method to symmetric objects without specifying symmetry axes,
whereas our method inherently optimizes to a suitable pose.
We also note that DeepIM is object-centric, refining each object's pose
separately. In contrast, our method retains the entire scene, refining all
object poses simultaneously and thus is able to account for inter-object
effects.

In this work, we use render-and-compare to refine 6D object poses in cluttered real-world scenes. Instead of comparing the rendered and observed RGB images, we propose to use an abstraction network to deal with the difficulties in comparing images from two different modalities.
\section{Learned Descriptors for Scene Abstraction}
\label{sec:descriptors}

Real scenes exhibit a large variety of secondary effects such as lighting,
camera noise, reflections, and so on. All of these effects are very difficult
to model and severely constrain the applicability of differentiable rendering
methods.
We propose an additional abstraction module $f: \mathbb{I} \rightarrow \mathbb{A}$,
mapping the RGB image space $I$ to an abstract feature space $A$. Ideally,
the mentioned secondary effects lie in the null space of $f$.
For convenience, we require that $A$ is also image-like so that pixels in
$I$ correspond to feature vectors in $A$.

The very difficult problem of decomposing an image into its different intrinsic components, such as shading, reflectance, and shape,
has been studied extensively \citep{barrow1978recovering,tappen2003recovering,finlayson2004intrinsic}.
However, in this application such a complex and physically accurate decomposition
is not required.
In order to be usable for differential rendering, we only require that features
are similar for corresponding points on the same object (under varying
lighting conditions etc), and dissimilar for non-corresponding objects.
Many traditional feature extractors exhibit this property (e.g. SIFT).
Of particular interest, however, are feature extractors designed for
dense output.
Recently, \cite{schmidt2017self} showed that highly precise
feature extractors can be trained in a self-supervised way from ground truth
correspondences. The learned descriptors outperform sparse feature extractors
by a large margin.

\subsection{Real-Synthetic Correspondences for Descriptor Learning}
\label{sec:rendering}

\begin{figure}
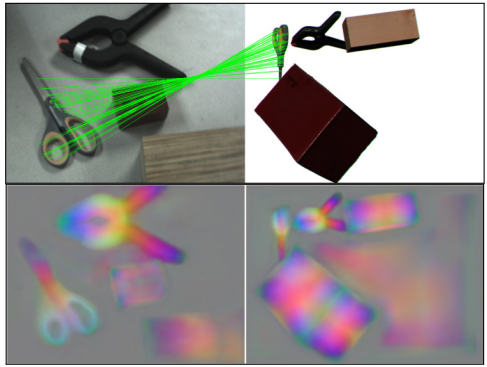

 \centering
 \includegraphics[height=2.38cm,frame]{figures/dense/dataset_train.png}
 \includegraphics[height=2.38cm,frame]{figures/dense/dataset_train_features.png}
 \caption{Learning dense descriptors from real-synthetic correspondences.
 Top: Real and synthetic input frames $A$ and $B$. Positive correspondences
 matches are shown in green for one object.
 Bottom: Learned dense abstract representation of the scene.}
 \label{fig:dense:train}
\end{figure}

In order to leverage this idea in the differential rendering setting, we propose
to learn descriptors from the object meshes in conjunction with a training
dataset for pose estimation. For a pose-annotated real dataset frame $A$,
we render a synthetic frame $B$ with the same object set, but using
different poses (see \cref{fig:dense:train}).
Corresponding points in both RGB frames can be easily determined
from the object poses through projective geometry.

With a probability of 0.5, the synthetic object is drawn in the same orientation
as the real object (see \cref{fig:dense:train}). This results in a large
number of positive correspondences. In the other case, the orientation is drawn
uniformly random---resulting in a larger number of negative examples and ensuring
that the learned descriptors stay globally unique.
The object translation is always sampled uniformly at random.

\subsection{Network Details \& Training}

Since our application depends on high spatial resolution of the computed
features, we follow the architecture of
Light-Weight~RefineNet~\cite{nekrasov2018light}, a
state-of-the-art semantic segmentation method, which successively upsamples
lower-resolution feature maps of higher abstraction levels and combines them
with higher-resolution feature maps of lower abstraction levels.
In our network, we use ResNet-34~\cite{he2016deep} with ImageNet pretraining as
the RefineNet backbone network.
The final convolutional layer is adapted to not only output semantic segmentation
($C{\times}H{\times}W$) but also the dense features ($D{\times}H{\times}W$).
Here, semantic segmentation is included as an auxiliary task, as it is not
used in the following pipeline stages (but could be in the future).
In our experiments, we use $D=3$ for easier visualization, as in
\cite{florence2018dense}.

Following \cite{schmidt2017self} and \cite{florence2018dense} we minimize a
pixel-wise contrastive loss function $\mathcal{L}_C(A,B)$:
\begin{align}
 \mathcal{L}_+(A,B) &= \frac{1}{|M_+|} \sum_{u \in M_+} || f_A(u_A) - f_B(u_B) ||_2^2 \\
 \mathcal{L}_-(A,B) &= \frac{1}{H_-} \sum_{u \in M_-} \max (0, M - || f_A(u_A) - f_B(u_B) ||_2)^2 \\
 \mathcal{L}_C(A,B)   &= \mathcal{L}_+(A,B) + \mathcal{L}_-(A,B),
\end{align}
where $f_P(u)$ is the descriptor value in image $P$ at location $u$,
$M_+$ is the set of correspondent pixel pairs $(u_A, u_B)$, $M_-$ is a set
of randomly sampled negative correspondences (also limited to the object mask),
and $H_- = \sum_{u \in M_-} \mathds{1}[M - || f_A(u_A) - f_B(u_B) ||_2 > 0]$ the number of
\textit{hard negatives}.

To encourage the network to disregard clutter in the background, we also
introduce a loss on background pixels for a single frame $F$:
\begin{align}
 \mathcal{L}_\text{bg}(F) &= \lambda \frac{1}{G} \sum_{u \in G} || f_F(u) ||_2^2,
\end{align}
where $G$ is the set of background pixels. The loss balancing factor $\lambda=0.1$
is chosen rather small in order not to hurt descriptor learning in the foreground
pixels.

The combined loss function is simply
\begin{align}
 \mathcal{L}(A,B) &= \mathcal{L}_C(A,B) + \sum_{F \in \{A,B\}} \mathcal{L}_S(F) + \mathcal{L}_\text{bg}(F),
\end{align}
where $\mathcal{L}_S$ is the cross-entropy loss for pixel-wise segmentation.

The network is trained using the Adam optimizer with learning rate 1e-4 and
parameters $\beta_1 = 0.9, \beta_2 = 0.999$ on 350k image pairs. Note that
the synthetic image $B$ of each image pair is generated on the fly, i.e. the
network never is presented with the same pair twice.

\subsection{Mesh Representation for Surface Features}
\label{sec:fusion}

\begin{figure}
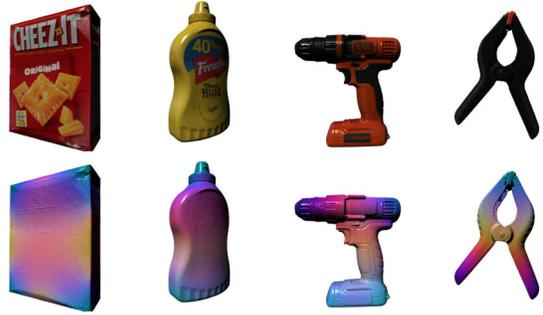

 \centering
 \newlength{\imgw}
 \setlength{\imgw}{2cm}
 \setlength{\tabcolsep}{0cm}
 \begin{tabular}{cccccc}
  \includegraphics[width=\imgw]{figures/dense/dataset_rgb/003_cracker_box.jpg} &
  \includegraphics[width=\imgw]{figures/dense/dataset_rgb/006_mustard_bottle.jpg} &
  \includegraphics[width=\imgw]{figures/dense/dataset_rgb/035_power_drill.jpg} &
  \includegraphics[width=\imgw]{figures/dense/dataset_rgb/051_large_clamp.jpg} \\

  \includegraphics[width=\imgw]{figures/dense/dataset_features/003_cracker_box.jpg} &
  \includegraphics[width=\imgw]{figures/dense/dataset_features/006_mustard_bottle.jpg} &
  \includegraphics[width=\imgw]{figures/dense/dataset_features/035_power_drill.jpg} &
  \includegraphics[width=\imgw]{figures/dense/dataset_features/051_large_clamp.jpg} \\
 \end{tabular}
 \caption{Learned surface features, projected and fused onto the mesh.
 The 3D feature vectors are visualized directly as RGB colors.}
 \label{fig:dense:fused_meshes}
\end{figure}

Since the learned features should be constant for each local surface patch,
independent of the viewing pose, we can \textit{fuse} the descriptor information
onto the mesh representation. To this end, we render $N$ views (50 in our experiments)
of the object
from randomly sampled viewing directions and viewing distances. After feature
extraction using the learned network, the resulting point-feature pairs are
aggregated in the object frame. A voxel grid downsampling is applied, where
descriptors and positions are each averaged inside each voxel.
The voxel size is determined heuristically from the object bounding box s.t.
the voxel count is constant---this results in constant-size output.
This method is robust and easy to tune in case more complex geometry needs to
be supported. In our experiments, we use 5000 voxels.
Finally, each vertex of the object mesh is assigned an interpolated descriptor
from the four nearest voxels using inverse-distance weighting.
\Cref{fig:dense:fused_meshes} shows exemplary meshes and corresponding feature
visualizations.

\section{Differentiable Renderer}

The pose refinement problem is an optimization problem. In our case, we assume
that we start with a pose initialization of reasonable quality, such that
local optimization methods can find the optimum solution.
In this context, it is very favorable to be able to compute derivatives
of the rendering process, since the number of parameters grows linearly with
the number of objects in the scene (at least six parameters per object).
Optimization without gradient information quickly becomes infeasibly slow.

We base our differential rendering module on the method of
OpenDR~\citep{loper2014opendr}, which is able to approximate gradients with respect
to lighting parameters, camera parameters, object poses, etc.
We note that in our setting, only pose parameters need to be optimized, because
lighting and other surface effects are removed by the abstraction network
and camera parameters are assumed to be fixed in monocular pose estimation.
Encouraged by this simplification, we implemented a lightweight differentiable
renderer, which we call \textit{LightDR}.

The OpenDR method is built around the screen-space approximation of the derivative
of the rendering process. Gradients due to occlusion effects during this 3D-2D
reduction are approximated from the local intensity gradient.
In essence, this idea assumes that occluded pixels are similar to their visible
neighbors.

In order to simplify gradient computation, we locally linearize the pose:
\begin{align}
T(\alpha,\beta,\gamma,a,b,c) &= T_0 \left( \begin{matrix}
   1 & -\gamma & \beta & a \\
   \gamma & 1 & -\alpha & b \\
   -\beta & \alpha & 1 & c \\
   0 & 0 & 0 & 1 \\
\end{matrix} \right)
\end{align}
The rotation part of $T$ is orthonormalized after each optimization step.

The scenes we are interested in feature high levels of occlusion between the
individual objects. \citep{loper2014opendr} makes several assumptions in computation
of the screen-space gradients.  While these assumptions help in simplifying the computation,
in real world scenarios, they are often violated.
We discuss the assumptions involved, scenarios where these assumptions are violated, and propose solutions
for better approximation of the pose gradients.

\subsection{Gradients on Occlusion Boundaries}
\begin{figure}
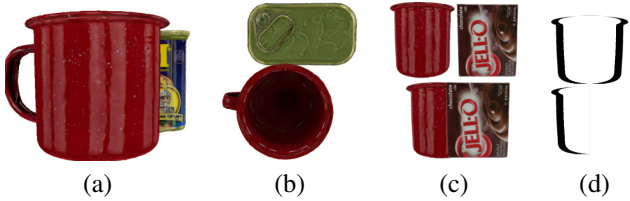

   \centering
   \newlength{\imgh}
   \setlength{\imgh}{2.1cm}
   \begin{tabular}{cccc}
   {\includegraphics[height=\imgh, clip, trim=10 10 0 0]{figures/dr/front_view.png}}&
   {\includegraphics[height=\imgh]{figures/dr/top_view.png}}&
   {\includegraphics[height=\imgh]{figures/dr/scene_top_bottom.png}}&
   {\includegraphics[height=\imgh, clip, trim=10 10 0 0]{figures/dr/grad_top_bottom.png}}\\
   (a)&(b)&(c)&(d)
   \end{tabular}
   \caption{Corner cases for render gradient estimation.
   (a) and (b): Front and top view of an exemplary scene with occlusion.
   (c) Top:  Occlusion-free scene; Bottom: Scene with mug occluded.
   (d) Magnitude of pixel-wise loss corresponding to scenarios depicted in (c)
   when the mug is translated. Black depicts: high loss magnitude.
   When the rendered mug is moved behind the occluder, all pixels with high
   loss (and thus gradient information) lie outside of the rendered object mask.
   }
   \label{fig:dr}
\end{figure}

Occlusion boundaries are highly important for pose refinement, since they
offer much information about the scene layout.
On the occlusion boundary pixels, OpenDR uses Sobel kernel ($\frac{1}{2} [-1, 0, 1 ]$) and its transpose to compute the gradient along the horizontal axis and the vertical axis respectively, with the underlying assumption that a shift in the occlusion boundary can be approximated by the replacement of the current pixel by the neighboring pixel (of the other object).
However, this assumption is valid only if the occlusion boundary pixel belongs to the object in the foreground. \cref{fig:dr} (a), (b) depict the front view and top view of an example scene where the mug is occluding the can. Translating the mug results in covering or uncovering of can pixels, which is well approximated using the local Sobel gradient.
Conversely, translating the can in the background does not result in covering/uncovering mug pixels, rather, more can pixels will become visible or become covered. Thus using the Sobel derivative is incorrect in this case.
To address this issue, we detect such cases using the Z-buffer during rendering
suppress the Sobel gradient on these pixels.
We note that the occluded pixel belongs to the same object in this case, so that
zero gradient should be a good approximation.

\subsection{Propagating Image-space Gradients to Object Coordinates}

While propagating the image-space gradients to the object coordinates, only gradient from the pixels belonging to the object needs to be propagated. The na\"ive way to do this is to mask the image-space gradient with rendered object mask. However, in certain situations this means we are ignoring exactly the pixels where a pixel-wise loss function generates high gradients, namely just outside of the rendered object boundary. \Cref{fig:dr}(c-d) illustrates this point.

To address this issue, we propose a dilation of the rendered object mask by one
pixel, in order to include gradient information directly outside of the object
boundary.

\subsection{Implementation}
LightDR uses OpenGL via the Magnum engine\footnote{\url{https://magnum.graphics/}}
for forward rendering. The gradient backpropagation is built on top of PyTorch
to facilitate faster computations on GPU. LightDR needs a few milliseconds for
forward rendering and around 50\,ms for backpropagation of gradients to object
poses.

\section{Pose Refinement}

\begin{figure}
   \centering
   \scalebox{0.95}{%
\begin{tikzpicture}[
	fmap/.style={rectangle,draw=black,fill=yellow!20},
    font=\small\sffamily,
    label/.style={anchor=north,align=center, text depth=0},
	axislabel/.style={font=\footnotesize,inner sep=1pt},
]

\node[inner sep=0] (input) at (-4,0) {\includegraphics[width=.95\linewidth]{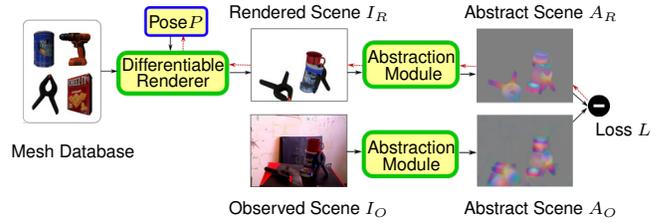}};

\node[label] at (-7.4, -.55) {\scriptsize Mesh Database};

\node[label] at (-4.1, 1.4) {\scriptsize  Rendered Scene $I_R$};
\node[label] at (-4.1, -1.35) {\scriptsize Observed Scene $I_O$};

\node[label] at (-0.85, 1.4) {\scriptsize Abstract Scene $A_R$};
\node[label] at (-0.85, -1.35) {\scriptsize  Abstract Scene $A_O$};

\node[label] at (-2.63, .775) {\scriptsize Abstraction };
\node[label] at (-2.6, .53) {\scriptsize Module };
\node[label] at (-2.63, -.5) {\scriptsize Abstraction };
\node[label] at (-2.63, -.73) {\scriptsize Module };

\node[label] at (-6., 0.66) {\scriptsize Differentiable};
\node[label] at (-6., 0.44) {\scriptsize Renderer};

\node[label] at (-5.96, 1.25) {\scriptsize Pose$P$};

\node[label] at (0.3, -0.25) {\scriptsize Loss $L$};
\end{tikzpicture}
}
   \caption{Render-Abstraction pipeline. The renderer produces an RGB image of
   the scene, which is then mapped into the abstract feature space.
   The loss gradient (red dotted line) is propagated back through the
   abstraction module and the differentiable renderer.}
   \label{fig:refinement:render_abstraction}
\end{figure}

Armed with the abstraction module and our differentiable renderer, we can tackle
the 6D pose refinement problem in cluttered real-world scenes.
We first experimented with the architecture depicted in \cref{fig:refinement:render_abstraction}.
The 3D scene with the objects in the current estimated pose $P$ is rendered to generate image $I_R$.
The abstraction module (see \cref{sec:descriptors}) is used to generate abstract representations $A_R$ and $A_O$ from the images $I_R$ and $I_O$.
The loss $L$ is computed as the pixel-wise loss between $A_R$ and $A_O$.
Finally, we can derive the gradient of $L$ with respect to the poses $P$:
\begin{align}
\frac{\partial L}{ \partial P} &=  \frac{\partial I_R}{ \partial P} \cdot \frac{\partial A_R}{ \partial I_R} \cdot \frac{\partial L}{ \partial A_R}
\end{align}

\begin{figure}[b]
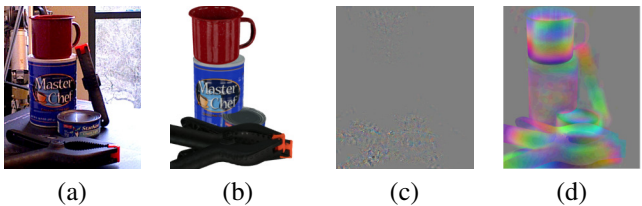

   \centering
   \newlength{\imgwdr}
   \setlength{\imgwdr}{2.15cm}
   \setlength{\tabcolsep}{0.2cm}
   \begin{tabular}{cccc}
    \includegraphics[height=\imgwdr,clip, trim=150 50 150 25]{figures/dr/grad/scene.png} &
    \includegraphics[height=\imgwdr,clip, trim=150 50 150 25]{figures/dr/grad/scene_rnd.png} &
    \includegraphics[height=\imgwdr,clip, trim=0 9 0 0]{figures/dr/grad/sparse.png} &
    \includegraphics[height=\imgwdr,clip, ]{figures/dr/grad/dense.png} \\
    (a)&(b)&(c)&(d)
   \end{tabular}
   \caption{Gradient of pixel-wise loss w.r.t. the rendered scene in the two pipeline variants.
   (a) Observed scene. (b) Rendered scene.
   (c) Render-Abstraction pipeline (see \cref{fig:refinement:render_abstraction}).
   (d) Abstraction-Render pipeline (see \cref{fig:refinement:abstraction_render}).
   The actual gradient magnitudes are scaled for better visualization.
   Gray corresponds to zero gradient.}
   \label{fig:dr:grad}
\end{figure}

As depicted in \cref{fig:refinement:render_abstraction},  $ \frac{\partial I_R}{ \partial P}$ is approximated by the differentiable renderer
and $\frac{\partial A_R}{\partial I_R}\cdot\frac{\partial L}{\partial A_R}$ is computed by standard backpropagation.
During experiments, we noticed that the latter gradient is rather sparse
and focuses most of its magnitude on few spatial locations in the image (see \cref{fig:dr:grad}).
This effect is well-known, for example in the field of adversarial example
generation for CNNs~\citep{goodfellow2015explaining}, \citep{palacio2018deep}. Here, it is highly undesirable,
since the differential rendering process works best with smooth, uniform
gradients.

\begin{figure*}
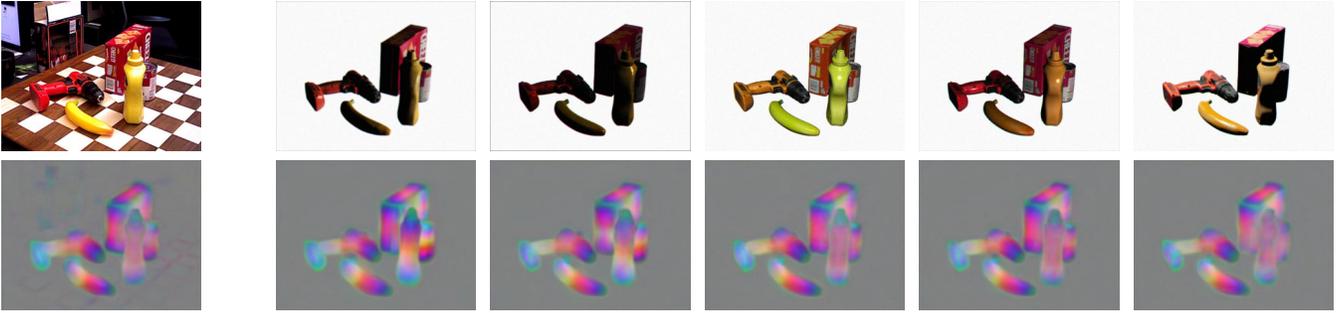

 \centering
 \newlength{\qwidth}
 \setlength{\qwidth}{2.65cm}
 \setlength{\tabcolsep}{0.1cm}
 \begin{tabular}{c@{\hspace{1cm}}ccccc}
  \includegraphics[width=\qwidth]{figures/dense/016/real.jpg} &
  \includegraphics[width=\qwidth]{figures/dense/016/ren_002.jpg} &
  \includegraphics[width=\qwidth]{figures/dense/016/ren_004.jpg} &
  \includegraphics[width=\qwidth]{figures/dense/016/ren_005.jpg} &
  \includegraphics[width=\qwidth]{figures/dense/016/ren_007.jpg} &
  \includegraphics[width=\qwidth]{figures/dense/016/ren_008.jpg} \\
  \includegraphics[width=\qwidth]{figures/dense/016/real_features.jpg} &
  \includegraphics[width=\qwidth]{figures/dense/016/ren_002_features.jpg} &
  \includegraphics[width=\qwidth]{figures/dense/016/ren_004_features.jpg} &
  \includegraphics[width=\qwidth]{figures/dense/016/ren_005_features.jpg} &
  \includegraphics[width=\qwidth]{figures/dense/016/ren_007_features.jpg} &
  \includegraphics[width=\qwidth]{figures/dense/016/ren_008_features.jpg} \\
 \end{tabular}
 \caption{Learned scene abstraction, removing secondary effects such as
   background, lighting, shadows, and camera noise.
   Top row: RGB input scene, bottom row: RGB visualization of the output feature
   descriptors.
   The leftmost column shows a real test scene from the YCB dataset, the other
   columns contain rendered scenes with random lighting and camera noise.}
 \label{fig:feature_quality}
\end{figure*}

\begin{figure*}
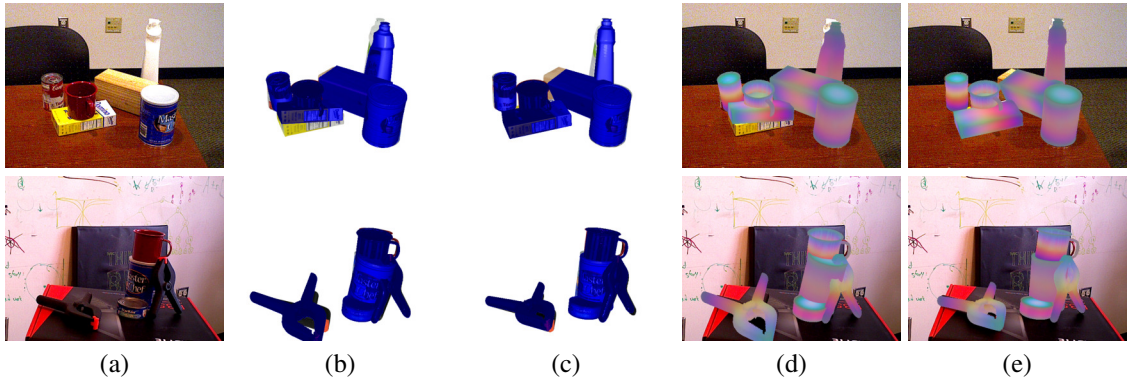

   \centering
   \newlength{\imgres}
   \setlength{\imgres}{2.9cm}
   \setlength{\tabcolsep}{0.05cm}
   \begin{tabular}{ccccc}
    \includegraphics[width=\imgres]{figures/results/000170_scene.png} &
    \includegraphics[width=\imgres]{figures/results/000170_init_rgb_000.png} &
    \includegraphics[width=\imgres]{figures/results/000170_opt_rgb_100.png} &
    \includegraphics[width=\imgres]{figures/results/000170_init_feat_000.png} &
    \includegraphics[width=\imgres]{figures/results/000170_opt_feat_100.png} \\
    \includegraphics[width=\imgres]{figures/results/000021_scene.png} &
    \includegraphics[width=\imgres]{figures/results/000021_init_rgb_000.png} &
    \includegraphics[width=\imgres]{figures/results/000021_opt_rgb_100.png} &
    \includegraphics[width=\imgres]{figures/results/000021_init_feat_000.png} &
    \includegraphics[width=\imgres]{figures/results/000021_opt_feat_100.png} \\
    (a) & (b) & (c) & (d) & (e)
   \end{tabular}
   \caption{Qualitative examples from the YCB Video Dataset.
   (a): Observed scene.
   (b) and (c): Renderings (blue) with initial and optimized pose parameters, respectively.
   (d) and (e): Renderings of the feature-annotated meshes in initial and
   optimized pose, respectively.}
   \label{fig:result}
\end{figure*}

To mitigate this issue, we investigated a second pipeline shown in \cref{fig:refinement:abstraction_render}.
We use the method described in \cref{sec:fusion} to create meshes with fused
surface descriptors.
Rendering these meshes directly results in the abstract rendered $A_R$.
In this case, $\frac{\partial L}{ \partial P} $ is simpler:
\begin{align}
\frac{\partial L}{ \partial P} &=  \frac{\partial L}{ \partial A_R} \cdot \frac{\partial A_R}{ \partial P}.
\end{align}
Here, $\frac{\partial A_R}{ \partial P}$ is approximated directly by the differentiable renderer.
An additional benefit of this variant is that only one forward pass of the abstraction module
for $A_O$ is required. This directly translates to significant reduction in the optimization process runtime.

We optimize the object poses with the AdaGrad optimization scheme. This avoids
manual tuning of learning rates for translation and rotation parameters---which
differ largely in scale. In our experiments, we use a learning rate $\lambda = 1e-2$
with a decay of $0.99$. The optimization runs for 50 iterations, which corresponds
to roughly 2\,s per frame.

\section{Experiments}

We perform our experiments on the YCB Video dataset~\citep{xiang2017posecnn},
which consists of 133,936 images extracted from 92 videos, showing 21 different
objects in cluttered arrangements.
Importantly, the dataset comes with high-quality meshes, which are also used for
synthetic data generation by most of the pose estimation methods applied to the
dataset \citep{xiang2017posecnn,oberweger2018making}.

We first qualitatively show the result of the learned mapping into the abstract
feature space in \cref{fig:feature_quality}. To be able to control
primary and secondary effects separately, we rely on our rendering pipeline
and generate multiple scenes with random secondary parameters as detailed in
\cref{sec:rendering}.
Our trained model is able to
effectively suppress the background pixels and produces robust, consistent
output under changing lighting conditions and camera model parameters.

For a quantitative analysis, we measure the
the ADD and ADD-S metrics as in \citep{xiang2017posecnn} for each
object occurrence, which measure average point-wise distances between transformed
objects and the ground truth, for non-symmetric and symmetric objects, respectively:

\begin{align}
\text{ADD} &= \frac{1}{m} \sum_{x\in \mathbb{M}} || (Rx + T) - (\tilde{R}x + \tilde{T})||, \\
\text{ADD-S} &= \frac{1}{m} \sum_{x_1\in \mathbb{M}}\min_{x_2\in\mathbb{M}} || (Rx_1 + T) - (\tilde{R}x_2 + \tilde{T})||,
\end{align}
where $R$ and $T$ are the ground truth rotation and translation,
$\tilde{R}$ and $\tilde{T}$ denote the estimated pose, and $\mathbb{M}$
is the set of model points as included in the YCB Video dataset.
We aggregate all results and measure the area under the
threshold-accuracy curve for distance thresholds from zero to 0.1\,m, which
is the same procedure as in \citep{xiang2017posecnn}.

\begin{table*}
 \centering
 \caption{Pose refinement results on the YCB Video Dataset}
 \label{tab:experiments:results_posecnn}
 \footnotesize\centering\setlength{\tabcolsep}{.1cm}
 \begin{filecontents*}{results/posecnn.txt}
Class initadd initadds refinedadd deltaadd refinedadds deltaadds                                                                                                   oinitadd oinitadds orefinedadd odeltaadd orefinedadds odeltaadds                                                            deepimadd deepimadds
master\_chef\_can                   0.501778244972229 0.8391937613487244 0.6327041983604431 0.1309259533882141 0.916751503944397 0.07755774259567261             0.8185297250747681 0.9139257669448853 0.7674334049224854 -0.051096320152282715 0.9018034934997559 -0.012122273445129395       0.652      0.878
cracker\_box                        0.5306612849235535 0.7686113715171814 0.6531434655189514 0.12248218059539795 0.8171980381011963 0.04858666658401489          0.8362063765525818 0.9002540111541748 0.8287955522537231 -0.007410824298858643 0.8942897319793701 -0.0059642791748046875      0.826      0.898
sugar\_box                          0.6841589212417603 0.8424947261810303 0.8533899784088135 0.16923105716705322 0.9203624129295349 0.07786768674850464          0.8211969137191772 0.8979843854904175 0.8641504049301147 0.0429534912109375 0.9222835302352905 0.024299144744873047           0.897      0.938
tomato\_soup\_can                   0.6615778207778931 0.8099872469902039 0.5935521721839905 -0.06802564859390259 0.7986630201339722 -0.01132422685623169        0.797607421875 0.8953881859779358 0.573995053768158 -0.22361236810684204 0.7821711897850037 -0.11321699619293213              0.814      0.901
mustard\_bottle                     0.8099198341369629 0.9043291807174683 0.8649253249168396 0.05500549077987671 0.9232683181762695 0.01893913745880127          0.9148505926132202 0.9500575661659241 0.8674660325050354 -0.047384560108184814 0.9258291721343994 -0.024228394031524658       0.903      0.944
tuna\_fish\_can                     0.7065482139587402 0.8804519176483154 0.8108333349227905 0.10428512096405029 0.9432834386825562 0.06283152103424072          0.48723873496055603 0.7167922258377075 0.6972867250442505 0.21004799008369446 0.857149600982666 0.1403573751449585            0.854      0.945
pudding\_box                        0.6269702911376953 0.7905269861221313 0.7113277912139893 0.08435750007629395 0.8313415050506592 0.04081451892852783          0.9017752408981323 0.9406352043151855 0.6876444816589355 -0.21413075923919678 0.8067503571510315 -0.13388484716415405         0.849      0.918
gelatin\_box                        0.7520696520805359 0.8721632361412048 0.8149638772010803 0.06289422512054443 0.8914321064949036 0.01926887035369873          0.9371086359024048 0.9585232138633728 0.7300469279289246 -0.20706170797348022 0.8275331258773804 -0.13099008798599243         0.877      0.916
potted\_meat\_can                   0.5950432419776917 0.7850133180618286 0.6373919248580933 0.04234868288040161 0.8033443689346313 0.018331050872802734         0.791067361831665 0.9000909328460693 0.7462961673736572 -0.04477119445800781 0.876233696937561 -0.0238572359085083            0.70       0.782
banana                              0.7233172059059143 0.8598577380180359 0.8208462595939636 0.09752905368804932 0.9181644916534424 0.058306753635406494         0.5169236660003662 0.6782258152961731 0.6875652074813843 0.17064154148101807 0.8101105690002441 0.13188475370407104           0.833      0.92
pitcher\_base                       0.5326846241950989 0.7698318958282471 0.8506777882575989 0.3179931640625 0.9265830516815186 0.15675115585327148              0.6944947242736816 0.8497997522354126 0.8381680250167847 0.14367330074310303 0.9205660820007324 0.07076632976531982           0.887      0.937
bleach\_cleanser                    0.5031413435935974 0.7155749797821045 0.6500656604766846 0.14692431688308716 0.8044672012329102 0.08889222145080566          0.7621984481811523 0.8545255661010742 0.7825958728790283 0.020397424697875977 0.8763495087623596 0.0218239426612854           0.759      0.868
bowl                                0.033330287784338 0.6961195468902588 0.06462591886520386 0.03129563108086586 0.7554620504379272 0.05934250354766846          0.035815801471471786 0.7807824611663818 0.015171466395258904 -0.020644335076212883 0.6643458604812622 -0.11643660068511963    0.415      0.778
mug                                 0.5854103565216064 0.7815499305725098 0.6591536998748779 0.07374334335327148 0.8404676914215088 0.05891776084899902          0.5390497446060181 0.7576519250869751 0.578779935836792 0.039730191230773926 0.7888040542602539 0.03115212917327881           0.703      0.861
power\_drill                        0.5530976057052612 0.7269642353057861 0.7372171878814697 0.1841195821762085 0.8589481711387634 0.1319839358329773            0.8287023305892944 0.9081723690032959 0.8154937624931335 -0.013208568096160889 0.9044217467308044 -0.003750622272491455       0.907      0.946
wood\_block                         0.26612016558647156 0.6433285474777222 0.4546429216861725 0.18852275609970093 0.7331782579421997 0.08984971046447754         0.0 0.570010781288147 0.0 0.0 0.6033506989479065 0.03333991765975952                                                          0.262      0.601
scissors                            0.35818105936050415 0.568753182888031 0.39964523911476135 0.0414641797542572 0.5861743688583374 0.017421185970306396         0.652976393699646 0.7963002324104309 0.7544541358947754 0.1014777421951294 0.854475200176239 0.058174967765808105             0.455      0.618
large\_marker                       0.5826793909072876 0.7172664403915405 0.639059841632843 0.05638045072555542 0.7734977602958679 0.05623131990432739           0.5651437640190125 0.7018232345581055 0.5978606939315796 0.03271692991256714 0.701823890209198 6.556510925292969e-07          0.681      0.775
large\_clamp                        0.24591559171676636 0.5015565752983093 0.37022337317466736 0.124307781457901 0.6513856649398804 0.14982908964157104          0.5718809962272644 0.7310830950737 0.7532730102539062 0.18139201402664185 0.8564221858978271 0.1253390908241272               0.455      0.721
extra\_large\_clamp                 0.16055412590503693 0.4411022663116455 0.25373655557632446 0.09318242967128754 0.6373697519302368 0.1962674856185913          0.23557528853416443 0.5460394024848938 0.20410476624965668 -0.03147052228450775 0.5831738114356995 0.037134408950805664      0.291      0.70
foam\_brick                         0.402414470911026 0.8799108862876892 0.43348100781440735 0.031066536903381348 0.9079477190971375 0.028036832809448242        0.32069313526153564 0.8889523148536682 0.370445191860199 0.04975205659866333 0.9208683967590332 0.03191608190536499           0.705      0.83
ALL 0.5370666980743408 0.7581213712692261 0.6278905272483826 0.09082382917404175 0.8242305517196655 0.06610918045043945                                            0.6624900102615356 0.8242332339286804 0.6698488593101501  0.007358849048614502 0.8349334597587585 0.010700225830078125      0.701      0.842
\end{filecontents*}
\pgfplotstabletypeset[
	every column/.style={
		column type=r,
		/pgf/number format/fixed,
		/pgf/number format/fixed zerofill,
		/pgf/number format/precision=1,
	},
	columns/Class/.style={
		column name={Object},
		column type=l,
		string type,
		string replace={SymC}{Class-wise average},
	},
	columns/initadd/.style={
		column name={ADD},
		multiply with=100,
	},
	columns/initadds/.style={
		column name={ADD-S},
		multiply with=100,
	},
	columns/refinedadd/.style={
		column name={ADD},
		multiply with=100,
	},
	columns/refinedadds/.style={
		column name={ADD-S},
		multiply with=100,
	},
	columns/deltaadd/.style={
		column name={$\Delta$},
		column type={@{(}r@{)\hspace{0.04cm}}},
		multiply with=100,
		showpos,
		fonts by sign={\color{green!50!black}}{\color{red}},
	},
	columns/deltaadds/.style={
		column name={$\Delta$},
		column type={@{(}r@{)\hspace{0.04cm}}},
		multiply with=100,
		showpos,
		fonts by sign={\color{green!50!black}}{\color{red}},
	},
	columns/oinitadd/.style={
		column name={ADD},
		column type={@{\hspace{.4cm}}r},
		multiply with=100,
	},
	columns/oinitadds/.style={
		column name={ADD-S},
		multiply with=100,
	},
	columns/orefinedadd/.style={
		column name={ADD},
		multiply with=100,
	},
	columns/orefinedadds/.style={
		column name={ADD-S},
		multiply with=100,
	},
	columns/odeltaadd/.style={
		column name={$\Delta$},
		column type={@{(}r@{)\hspace{0.04cm}}},
		multiply with=100,
		showpos,
		fonts by sign={\color{green!50!black}}{\color{red}},
	},
	columns/odeltaadds/.style={
		column name={$\Delta$},
		column type={@{(}r@{)\hspace{0.04cm}}},
		multiply with=100,
		showpos,
		fonts by sign={\color{green!50!black}}{\color{red}},
	},
	columns/deepimadd/.style={
		column name={ADD},
		column type={@{\hspace{.4cm}}r},
		multiply with=100,
	},
	columns/deepimadds/.style={
		column name={ADD-S},
		column type={r},
		multiply with=100,
	},
	every head row/.style={
		before row={\toprule
		 & \multicolumn{2}{c}{PoseCNN~\citep{xiang2017posecnn}} & \multicolumn{4}{c}{PoseCNN refined (ours)} & \multicolumn{2}{c}{HeatMaps~\citep{oberweger2018making}} & \multicolumn{4}{@{}c@{}}{HeatMaps refined (ours)} & \multicolumn{2}{c}{DeepIM~\citep{li2018deepim}} \\
		 \cmidrule (lr) {2-3} \cmidrule (lr) {2-3} \cmidrule (lr{.5cm}) {4-7} \cmidrule (lr) {8-9} \cmidrule (lr{.5cm}) {10-13} \cmidrule (lr) {14-15}
		},
		after row={\midrule},
	},
	every last row/.style={after row=\bottomrule},
	every row no 21/.style={before row=\midrule},
 ]
 {results/posecnn.txt}

 \vspace{0.1cm}
 We report the area
 under the accuracy curve (AUC) for varying error thresholds on the ADD and ADD-S metrics.
\end{table*}

We demonstrate pose refinement from the initialization of
PoseCNN~\citep{xiang2017posecnn} and the newer method by \citet{oberweger2018making}.
\Cref{fig:result} displays qualitative refinement examples, while
\Cref{tab:experiments:results_posecnn} gives quantitative results.
In our experiments, we assume that objects were correctly detected so that
we can focus on the problem of refining poses rather than correcting detections.
Our pipeline gives consistent improvements across nearly all objects of the
dataset for the PoseCNN initialization.
Note that we do not compare against the PoseCNN variant with ICP
post-refinement, since our pipeline works with RGB only and ICP requires
depth measurement.
The improvement is especially significant for large and textured objects.
On the initializations of \citet{oberweger2018making}, which are already of
very high quality, our gains are smaller.
We hypothesize that our approach is currently limited by the spatial resolution
of the computed feature representation.

Finally, compared to DeepIM~\cite{li2018deepim}, our method almost reaches
the same overall performance. We note that the experiments performed in
\citep{li2018deepim} apparently started from a better PoseCNN initialization
than what was available to us, though the difference seems small.
Interestingly, our method obtains significantly better results on a few object
classes---suggesting that a combination of the techniques (e.g. by
making the abstract representation and computed pose updates accessible to the
DeepIM network) could yield further improvements.

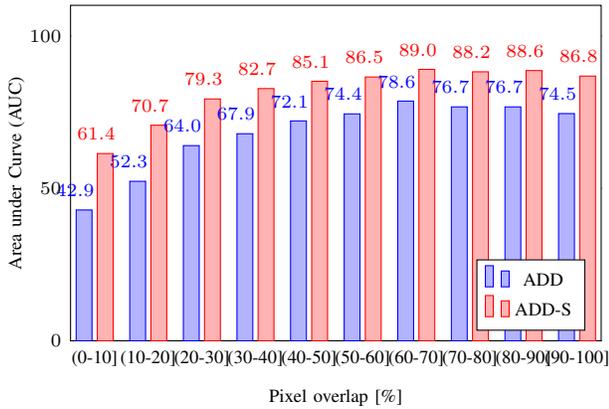
\begin{figure}
   \centering
\begin{tikzpicture}
   \begin{axis}[
     ybar,
     ymin=0, ymax=110,
     bar width=6pt,
     font=\scriptsize,
     tickwidth         = 0pt,
     enlarge x limits  = 0.05,
     xtick={{(0-10]}, {(10-20]},{(20-30]}, {(30-40]}, {(40-50]}, {(50-60]}, {(60-70]}, {(70-80]}, {(80-90]}, {(90-100]} },
     xtick style={draw=none},
     symbolic x coords = {{(0-10]}, {(10-20]},{(20-30]}, {(30-40]}, {(40-50]}, {(50-60]}, {(60-70]}, {(70-80]}, {(80-90]}, {(90-100]} },
     nodes near coords,
 legend pos=south east,
    ylabel={Area under Curve (AUC)},
    xlabel={Pixel overlap [\%]},
    width=\linewidth,
    height=.7\linewidth,
    nodes near coords style={anchor=south east,inner sep=0,shift={(4pt, 5pt)},font=\scriptsize,/pgf/number format/.cd,fixed,fixed zerofill,precision=1},
    ylabel near ticks, ylabel shift={-5pt},
   ]
   \addplot coordinates { 
 ({(0-10]},     42.9)
 ({(10-20]},    52.3)
 ({(20-30]},    64.0)
 ({(30-40]},    67.9)
 ({(40-50]},    72.1)
 ({(50-60]},    74.4)
 ({(60-70]},    78.6)
 ({(70-80]},    76.7)
 ({(80-90]},    76.7)
 ({(90-100]},   74.5)};
 
   \addplot coordinates { 
 
 ({(0-10]},     61.4)
 ({(10-20]},    70.7)
 ({(20-30]},    79.3)
 ({(30-40]},    82.7)
 ({(40-50]},    85.1)
 ({(50-60]},    86.5)
 ({(60-70]},    89.0)
 ({(70-80]},    88.2)
 ({(80-90]},    88.6)
 ({(90-100]},   86.8)};
   \legend{ADD, ADD-S}
   \end{axis}
 \end{tikzpicture}
    \caption{Basin of attraction in translation dimensions. We show resulting \mbox{ADD/ADD-S}
   metrics for varying initial 2D overlap of ground truth pose and initial
   estimate.}
   \label{fig:basin_trans}
\end{figure}

\begin{figure}
   \centering
\begin{tikzpicture}
   \begin{axis}[
     ybar,
     ymin=0, ymax=110,
     bar width=6pt,
     font=\scriptsize,
     tickwidth         = 0pt,
     enlarge x limits  = 0.05,
     xtick={{(0-5]}, {(5-10]},{(10-15]}, {(15-20]}, {(20-25]}, {(25-30]}, {(30-35]}, {(35-40]}, {(40-45]} },
     xtick style={draw=none},
     symbolic x coords = {{(0-5]}, {(5-10]},{(10-15]}, {(15-20]}, {(20-25]}, {(25-30]}, {(30-35]}, {(35-40]}, {(40-45]} },
     nodes near coords,
 legend pos=south west,
    ylabel={Area under Curve (AUC)},
    xlabel={Angular perturbation [$^\circ$]},
    width=\linewidth,
    height=.7\linewidth,
    nodes near coords style={anchor=south east,inner sep=0,shift={(4pt, 5pt)},font=\scriptsize,/pgf/number format/.cd,fixed,fixed zerofill,precision=1},
    ylabel near ticks, ylabel shift={-5pt},
   ]
   \addplot coordinates { 
 ({(0-5]},      90.3)
 ({(5-10]},     89.1)
 ({(10-15]},    83.7)
 ({(15-20]},    82.7)
 ({(20-25]},    77.5)
 ({(25-30]},    74.3)
 ({(30-35]},    69.2)
 ({(35-40]},    62.7)
 ({(40-45]},    58.1)};
 
   \addplot coordinates { 
 ({(0-5]},      94.5)
 ({(5-10]},     94.4)
 ({(10-15]},    92.0)
 ({(15-20]},    91.8)
 ({(20-25]},    89.6)
 ({(25-30]},    88.4)
 ({(30-35]},    86.1)
 ({(35-40]},    83.1)
 ({(40-45]},    80.8)};
   \legend{ADD, ADD-S}
   \end{axis}
 \end{tikzpicture}
    \caption{Basin of attraction in rotation dimensions. We show resulting \mbox{ADD/ADD-S}
   metrics for varying initial angular perturpations from the ground truth pose.}
   \label{fig:basin_rot}
\end{figure}
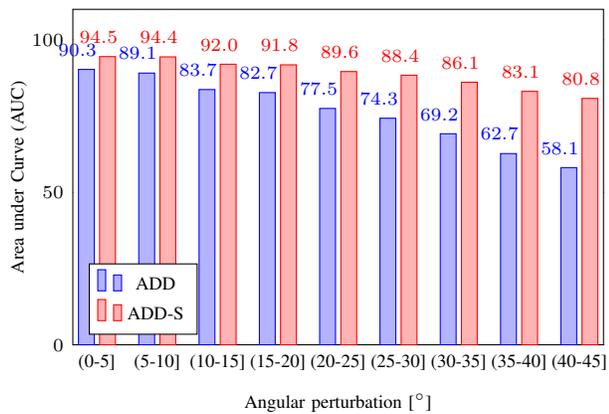

To quantify the robustness of our render-and-compare pipeline to the quality of the initialization, we analyzed
the basin of attraction of the refinement process. We experimented with 295 scenes from the validation set of 
YCB-Video dataset ($\sim$10 \% of the total validation scenes) by randomly perturbing the translation and
rotation components of the ground truth poses to varying degrees and optimizing the perturbed poses.
The translation perturbations were uniformly sampled in a range of $ \pm 5$ centimeters. 
Since the impact the translation perturbations has for an object depends of the size of the object,
we compute the percentage of pixel overlap between the observed image and the rendered image for an object.

Similarly, we uniformly sample an axis of rotation and an rotation angle in the range $ \pm 45$ degrees.
The AUC of the optimized pose with respect to different overlaps is shown in \cref{fig:basin_trans} and the rotation angle
is shown in \cref{fig:basin_rot}.
Our method is able to robustly handle translation perturpations with almost no
loss in accuracy down to $30\%$ remaining overlap. In the rotation experiment,
the ADD-S metric is almost unaffected by rotations of up to $45^\circ$.
The ADD metric drops off more steeply---this is caused by the entirely symmetric
objects, where the system has no chance of correcting the perturbation around
the symmetry axis.

\section{Conclusion}

We introduced a technique for scene abstraction by dense object features
learned in a self-supervised way and scene analysis by render-and-compare
utilizing a fast differential renderer implementation.
Our proposed method yields good results on the challenging YCB Video dataset,
where it robustly refines rough initial pose estimates to precise localizations.
We further demonstrated its large basin of attraction from perturbed initializations.
We see our result as a proof-of-concept for differential rendering in the
context of scene analysis.
In future work, we will increase performance further
and investigate other applications, for example non-rigid registration of
category-level models.
Due to its formulation the method could also be combined with other
iterative refinement procedures, contributing its holistic scene understanding.

\printbibliography

\end{document}